\documentclass[10pt,twocolumn,letterpaper]{article}

\usepackage{wacv}              % To produce the CAMERA-READY version
\usepackage{xcolor}
\usepackage{multirow}
\usepackage[T1]{fontenc}
\definecolor{wacvblue}{rgb}{0.21,0.49,0.74}
\usepackage[pagebackref,breaklinks,colorlinks,allcolors=wacvblue]{hyperref}

\def\wacvPaperID{2642} % *** Enter the WACV Paper ID here
\def\confName{WACV}
\def\confYear{2026}

\title{VOCAL: Visual Odometry via ContrAstive Learning}

\author{
Chi-Yao Huang\thanks{These authors contributed equally to this work.} \quad
Zeel Bhatt\footnotemark[1] \quad
Yezhou Yang\\[2mm]
Arizona State University\\
{\tt\small \{cy.huang, zbhatt1, yz.yang\}@asu.edu}
}

\begin{document}
\maketitle

\begin{abstract}
Breakthroughs in visual odometry (VO) have fundamentally reshaped the landscape of robotics, enabling ultra-precise camera state estimation that is crucial for modern autonomous systems. Despite these advances, many learning-based VO techniques rely on rigid geometric assumptions, which often fall short in interpretability and lack a solid theoretical basis within fully data-driven frameworks. To overcome these limitations, we introduce \textbf{VOCAL} (Visual Odometry via ContrAstive Learning), a novel framework that reimagines VO as a label ranking challenge. By integrating Bayesian inference with a representation learning framework, VOCAL organizes visual features to mirror camera states. The ranking mechanism compels similar camera states to converge into consistent and spatially coherent representations within the latent space. This strategic alignment not only bolsters the interpretability of the learned features but also ensures compatibility with multimodal data sources. Extensive evaluations on the KITTI dataset highlight VOCAL’s enhanced interpretability and flexibility, pushing VO toward more general and explainable spatial intelligence.
\end{abstract}    
\section{Introduction} \label{sec:intro}

\begin{figure}[t] \centering \includegraphics[width=1.0\linewidth]{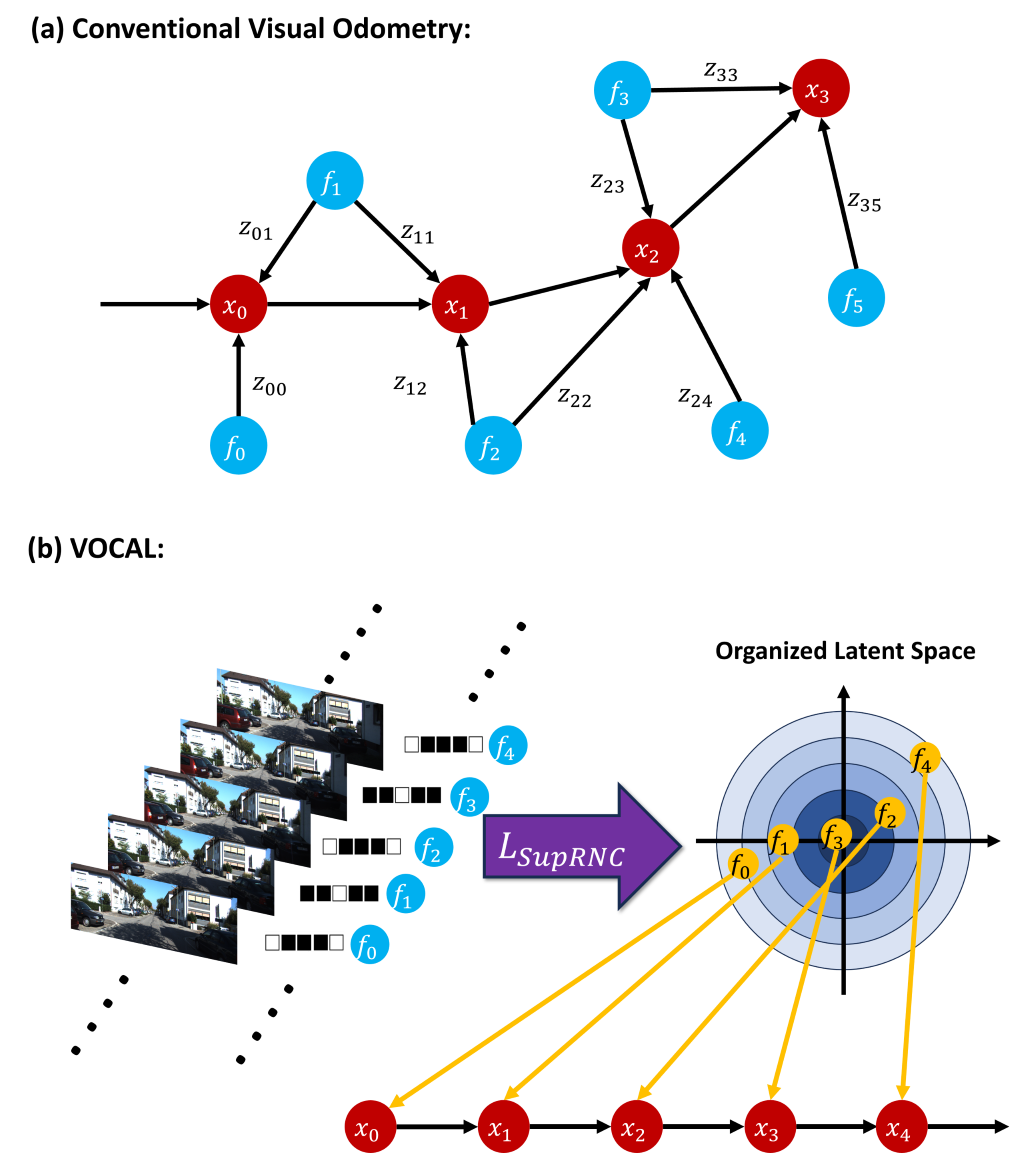} \caption{ \textbf{(a)} Conventional graph-based visual odometry, where connections between camera states \(x_i\) and features \(f_j\) are modeled using predefined graphs \(z_{ij}\). This manual design limits both flexibility and interpretability in learning-based VO systems.
\textbf{(b)} The VOCAL architecture eliminates the need for handcrafted graph structures by reframing VO as a label-ranking problem. Through contrastive learning, VOCAL organizes features extracted from visual inputs based on their corresponding camera states, ensuring that inputs with similar camera states yield consistent features in the latent space. Our approach improves spatial understanding in visual odometry by establishing a direct correlation between feature representations and 3D camera states. } \label{vocal_main} \end{figure}

As artificial intelligence advances, the need for multimodal learning methods continues to grow. Visual Odometry (VO)—a key technology for motion estimation in robotics, augmented reality, and autonomous driving—has traditionally relied on geometric constraints, temporal consistency, handcrafted features, and bundle adjustment. While these classical methods are effective, they were developed outside the deep learning paradigm, which instead relies on learning from latent representations.

This divergence creates a notable gap: conventional VO systems struggle to integrate with learning-based frameworks that operate in latent space. Although some learning-based VO models have been proposed, many still depend on geometric constraints that are difficult to interpret and not well aligned with representation learning. As a result, they are hard to adapt to broader multimodal systems such as large language models (LLMs) \cite{GPT4-V} and vision-language models (VLMs) \cite{LLaVA}. Closing this gap is critical for enabling VO to collaborate with other learning-based models and to advance spatial intelligence.

In recent years, contrastive learning has emerged as a powerful tool for representation learning. Originally developed as a self-supervised technique, it learns to group similar samples and separate dissimilar ones in the latent space. Methods such as SimCLR \cite{SimCLR} and MoCo \cite{MoCo} have demonstrated its effectiveness in image classification. Extensions like RankSim \cite{RankSim} and Rank-N-Contrast \cite{RNC} further apply ranking mechanisms in latent space to improve regression tasks. Similarly, \cite{PoseContrast} and \cite{Trajectory-Regularization} show that structured latent representations benefit object pose estimation. Contrastive learning has also become central to multimodal learning. For example, CLIP \cite{CLIP} aligns image and text pairs in a shared latent space using contrastive objectives. Large-scale multimodal systems such as \cite{GPT4-V}, Gemini \cite{gemini}, and LLaVA \cite{LLaVA} adopt similar principles to integrate diverse data types.

Despite these successes, the application of contrastive learning to visual odometry remains underexplored. The lack of organized and interpretable latent representations makes it difficult for VO systems to collaborate with other learning-based models.

To address the challenge, we propose VOCAL (Visual Odometry via ContrAstive Learning), a framework that applies contrastive learning to VO to produce structured, interpretable representations (see Fig.~\ref{vocal_main}). VOCAL reframes VO as a label-ranking problem, learning relationships between features and camera states in the latent space without relying on geometric constraints or handcrafted graph structures. This approach leads to a compact, explainable representation that integrates well with other learning-based systems and supports broader goals in spatial intelligence.

Our contributions are:
\begin{itemize}
\item We revisit Bayesian inference—the core principle behind VO—and reinterpret it in a modern latent representation framework.
\item We reformulate learning-based VO as a label-ranking problem, introducing a new, explainable way to organize visual features by camera states.
\item We provide a detailed analysis of how contrastive learning structures latent space for VO, offering new insights and improving the spatial understanding of learning-based models.
\end{itemize}
\section{Related Work} \label{sec:related_work}

Research in visual odometry (VO) can be broadly categorized into three main paradigms: geometry-based, hybrid, and learning-based approaches.

Geometry-based Visual Odometry:
These methods rely on geometric principles to estimate 3D structure and determine camera pose. They are typically divided into two categories: direct and feature-based methods. Direct methods (e.g., \cite{DSO, LSD-SLAM}) compute camera motion and scene structure directly from pixel intensities by comparing brightness across consecutive frames. Feature-based methods (e.g., \cite{PTAM, ORB-SLAM, VINSMONO}) extract and match keypoints or corners across frames to infer motion.
Despite their success, these approaches depend on handcrafted features and require manual processes to relate features to camera states. They also lack latent representations, limiting compatibility with learning-based models and hindering progress toward integration with modern AI systems.

Hybrid Visual Odometry:
Hybrid methods combine learning techniques with traditional geometry-based VO to improve robustness and incorporate semantic understanding. These systems typically consist of two components: a front-end and a back-end. In the front-end, conventional feature pipelines are enhanced with learned features, often extracted via CNNs to estimate depth \cite{CNN-SLAM, D3VO} or via object detectors to provide semantic cues \cite{semantic-SLAM, SLAM++, DSP-SLAM, FroDO, MaskFusion}. Scene segmentation methods \cite{mask-slam, Kimera} further enrich the map with structural information.
In the back-end, traditional optimization techniques like bundle adjustment are refined using learned priors \cite{GNets, Learning-to-BA}. However, hybrid systems still follow the geometric VO pipeline and require manual efforts to relate visual features to motion, limiting their interpretability and flexibility.

Learning-based Visual Odometry:
Purely learning-based VO methods aim to estimate camera pose using data-driven approaches. Early work like \cite{PoseNet} applied supervised learning to predict camera pose from RGB images, while \cite{DeepVO} used recurrent networks to model temporal dynamics. More recent methods such as \cite{UnDeepVO, SfMlearner} employ self-supervised learning using stereo image constraints, and \cite{TartanVO} introduces an intrinsics layer for improved generalization. \cite{Droid-SLAM} demonstrates a fully differentiable architecture that mimics geometric VO.
Despite these advances, many learning-based VO methods still incorporate geometric or temporal constraints and lack a clear and interpretable latent space, which makes integration with other learning-based systems difficult.

We draw inspiration from classical graph optimization tools \cite{GTSAM, g2o} and apply contrastive learning to address the limitations of existing VO methods. By organizing feature representations in a latent space and structuring their relationships with camera states through a label-ranking framework, our method enables a more interpretable and flexible alternative to traditional VO. This also promotes seamless integration with other learning-based systems, supporting broader applications in multimodal and cross-domain tasks.

\section{Methodology}
\label{sec:method}

We propose \textbf{VOCAL} (\textit{Visual Odometry via ContrAstive Learning}), a new framework that reformulates visual odometry (VO) by combining Bayesian inference with representation learning. In this section, we first outline the high-level motivation, then present a Bayesian view of VO. We follow this with a label-ranking formulation and describe how contrastive learning captures the relationship between visual features and camera states. This design aligns the latent space with 3D motion, improving spatial understanding and interpretability.

\subsection{High-Level Idea}
Our method is driven by two core goals. First, we aim to establish a consistent relationship between visual observations and camera states in latent space. Second, we seek to create an interpretable latent representation that can integrate seamlessly with other learning-based systems.

\begin{figure}[t] \centering \includegraphics[width=1.0\linewidth]{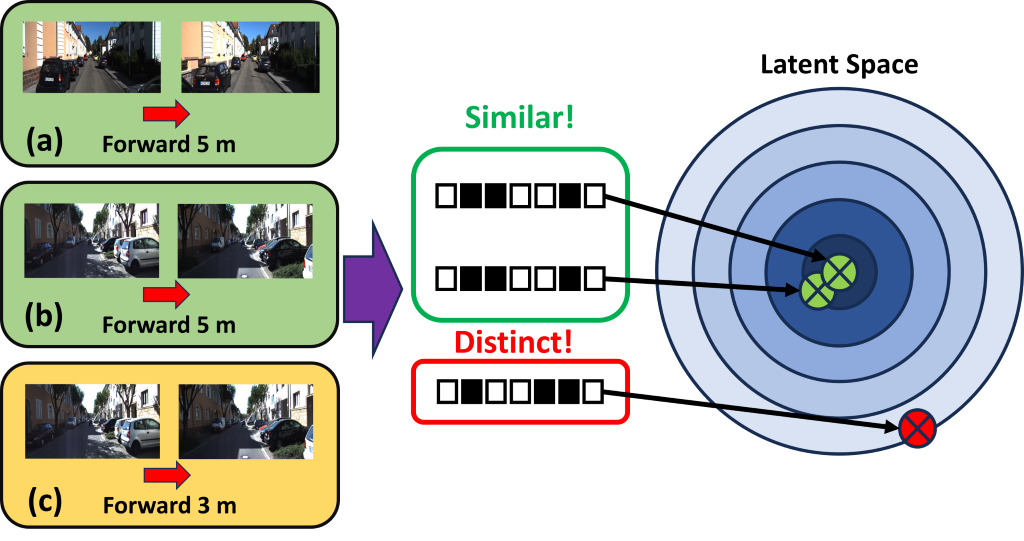} \caption{ \textbf{High-Level Idea:} Panels (a) and (b) show visual inputs from different environments that share the same camera state ("Forward 5 meters"), whereas panels (b) and (c) depict inputs from the same scene but with different camera states ("Forward 5 meters" vs. "Forward 3 meters"). Just as humans can recognize the same motion regardless of environmental differences—and distinguish different motions even in similar scenarios—our approach uses contrastive learning to align features corresponding to similar camera states while separating those corresponding to different states. } \label{vocal_idea} \end{figure}

Inspired by the human ability to recognize similar motion across diverse visual environments, we hypothesize that \textbf{visual inputs associated with similar camera states should be mapped to similar features in latent space—even if they originate from entirely different scenes} (Fig.~\ref{vocal_idea}).

To realize this, we adopt contrastive learning, which pulls similar samples closer and pushes dissimilar ones apart in latent space. By encouraging features from similar camera states to cluster, contrastive learning enables a flexible and spatially aligned representation that supports generalization and integration.

\subsection{Bayesian Inference in Learning-based VO} 
We begin by examining the VO problem through a Bayesian lens. Traditionally, VO is posed as a conditional probability estimation. Let
\[
    X = \{x_1, x_2, \ldots, x_N\}
\]
denote a set of camera states and 
\[
    Z = \{z_1, z_2, \ldots, z_N\}
\]
represent the corresponding observations, where \(N\) is the batch size. By Bayes’ rule, VO can be expressed as:
\begin{equation} 
    P(X \mid Z) \propto P(Z \mid X)\,P(X).
    \label{vo} 
\end{equation}

Typically, \(P(\cdot)\) is assumed to follow a Gaussian distribution; however, directly maximizing \(P(X \mid Z)\) is challenging. As a workaround, many methods approximate \(P(Z \mid X)\) via reprojection error or pixel intensity constraints under Gaussian assumptions.

Though these methods perform well in bundle-adjustment-based optimizers, applying a Gaussian model directly to learning-based VO is problematic because optimizing model parameters during training hinders direct parameter modeling in Eq.~\eqref{vo}. To address this issue, we reinterpret the training process by drawing parallels with conventional localization and mapping: we view backpropagation as inverting the observation model \cite{ProRobot} and treat training as a mapping function:
\begin{equation} 
    P(m_{\theta} \mid X, Z) \propto P(Z \mid X, m_{\theta})\,P(m_{\theta} \mid X),
    \label{new_vo_training} 
\end{equation}
where \(m_{\theta}\) denotes the learned model parameters that define the latent space. During inference, VO is reformulated as:
\begin{equation} 
    P(X \mid Z, m_{\theta}) \propto P(Z \mid X, m_{\theta})\,P(X \mid m_{\theta}).
    \label{new_vo_inference} 
\end{equation}

Despite this reformulation, estimating \(P(Z \mid X, m_{\theta})\) remains challenging, as the Gaussian assumption does not always align with the requirements of learning-based VO. This discrepancy has led many methods to rely on geometric loss functions that do not naturally fit within a purely probabilistic framework, underscoring the need for a more theoretically grounded approach.

\subsection{Label Ranking in Learning-based VO}
To overcome the limitation, we reformulate learning-based VO as a \emph{label-ranking} problem. In this formulation, camera states \(X\) serve as query instances, and observations \(Z\) are treated as labels. We adopt the Plackett–Luce model \cite{PL} to rank observations according to their corresponding camera states, providing an interpretable framework for organizing feature representations.

We define each ranked feature as:
\begin{equation} 
    f_{z_i} = m_{\theta}(X,\, z_i),
    \label{label ranking}
\end{equation}
where \(f_{z_i}\) denotes the feature associated with camera state \(x_i\), and \(m_{\theta}\) is the mapping function. Here, we assume that \(f_{z_i}\) is strictly positive. Under the Plackett–Luce model, the probability of a particular ranking order is given by:
\begin{equation} 
    P(z_1 > z_2 > \dots > z_N \mid X) = \prod_{k=1}^{N} \frac{f_{z_k}}{\sum_{j=k}^{N} f_{z_j}}.
    \label{label ranking probability}
\end{equation}

To solve this ranking problem, we employ the \emph{Rank-N-Contrast} (RNC) loss \cite{RNC}, which is specifically designed for continuous label regression tasks. Instead of treating each data point as a distinct class (as in contrastive learning for classification), RNC constructs positive/negative sample relationships based on the ordering of their queries, creating a ranked, continuous latent distribution that facilitates the estimation of the most likely camera states. The probability of each observation in the ranking is defined as:

\begin{equation}
    P\bigl(f_{z_j} \mid f_{z_i}, S_{i,j}\bigr) = \frac{\exp\Bigl(sim(f_{z_i}, f_{z_j})/\tau\Bigr)}{\sum_{f_{z_k} \in S_{i,j}} \exp\Bigl(sim(f_{z_i}, f_{z_k})/\tau\Bigr)},
    \label{rnc}
\end{equation}
where \(sim(\cdot,\cdot)\) is a similarity function, \(d(\cdot,\cdot)\) denotes the distance between camera states, and 
\[
    S_{i,j} := \{\,f_{z_k} \mid k \neq i,\ d(x_i,x_k) \ge d(x_i,x_j)\}
\]
is the set of samples ranked higher than \(f_{z_j}\), defining the ordering of camera states. The temperature \(\tau\) scales the similarity distribution to ensure stable training.

\begin{figure}[t]
  \centering
  \includegraphics[width=1.0\linewidth]{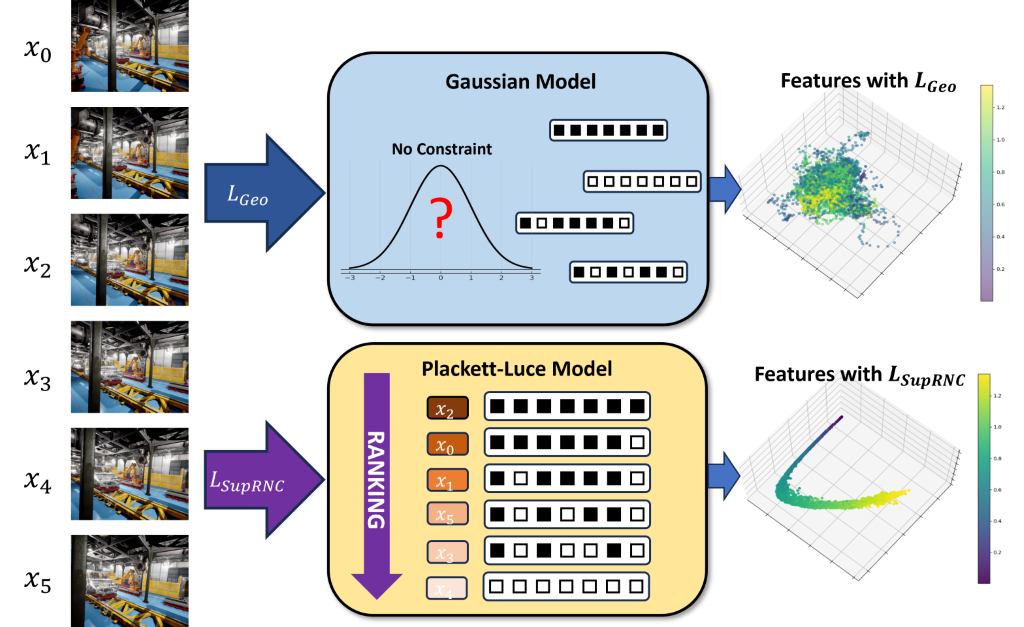}
  \caption{
    \textbf{Gaussian Model vs.\ Plackett–Luce Model in Learning-based VO:} Most learning-based VO methods rely on geometric loss functions derived from a Gaussian assumption, limiting their alignment with the learning process. In contrast, VOCAL adopts the Plackett–Luce model and employs the \emph{Supervised Rank-N-Contrast} loss (\(L_{SupRNC}\)) loss to rank feature representations according to their respective camera states, providing greater flexibility and a clearer interpretation of spatial relationships.
  }
  \label{vocal_label_ranking}
\end{figure}

Through contrastive learning, the model is encouraged to bring feature pairs \((f_{z_i}, f_{z_j})\) closer when their corresponding camera states \((x_i, x_j)\) are similar, and to push apart those that are dissimilar (see Fig.~\ref{vocal_label_ranking}). Formally,
\begin{equation}
    sim\bigl(f_{z_i}, f_{z_j}\bigr) \propto \frac{1}{d(x_i, x_j)}.
\end{equation}
Since \(f_{z_j}\) is obtained through this ranking process, it serves as a maximum likelihood estimate for the corresponding observation:
\begin{equation}
    P\bigl(f_{z_j} \mid f_{z_i}, S_{i,j}\bigr) \propto P\bigl(z_j \mid X, m_{\theta}\bigr).
\end{equation}
The ranking-based probability model is linked to learning-based VO, enabling VOCAL to organize observations by their camera states.

\begin{figure*}[t]
\centering
\includegraphics[width=0.9\linewidth]{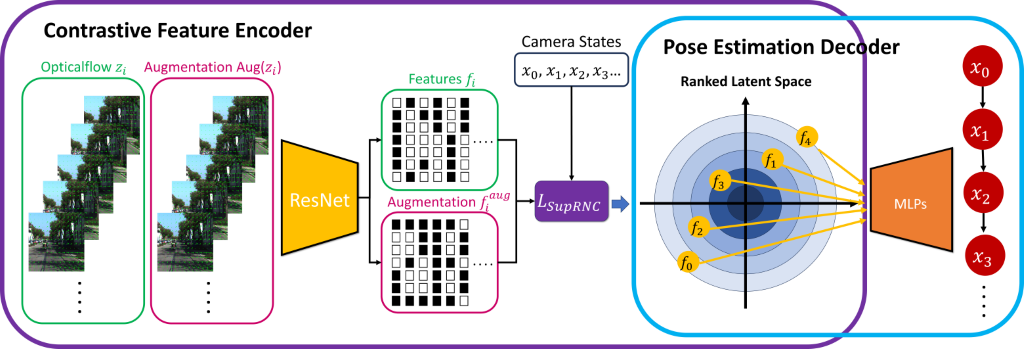}
\caption{
\textbf{System Overview:} Our system comprises two main components: a Contrastive Feature Encoder and a Pose Estimation Decoder. The encoder processes optical flow and its augmented variants using a ResNet to generate observation feature vectors. These features are then fed into the Pose Estimation Decoder, which employs Multi-Layer Perceptrons (MLPs) to estimate camera states. During training, the \emph{Supervised Rank-N-Contrast} loss (\(L_{SupRNC}\)) ranks the features based on camera states, yielding a spatially meaningful and interpretable latent space that facilitates the estimation of the most likely camera states.}
\label{vocal_overview}
\end{figure*}

\subsection{Visual Odometry via Contrastive Learning}
Having reformulated VO as a label-ranking task (Eq.~\eqref{label ranking}), we now describe how contrastive learning is incorporated into our framework. In our setup, each observation \(z_i\) consists of optical flow between two consecutive images, and each camera state \(x_i\) is represented by the corresponding relative camera motion. Our objective is to extract discriminative features from these observations and rank them according to their associated camera states.

We implement contrastive learning within the mapping function \(m_{\theta}\), which takes as input \(z_i\) along with its augmentation \(\mathrm{aug}(z_i)\). By organizing camera states into a ranking set \(S_{i,j}\), the label-ranking problem is reformulated as:
\begin{equation}
    m_{\theta}\Bigl(S_{i,j},\, \{z_i,\, \mathrm{aug}(z_i)\}\Bigr) \longrightarrow f_{z_i},\, f^{\mathrm{aug}}_{z_i}.
    \label{label ranking reformulated}
\end{equation}

We then apply the \emph{Rank-N-Contrast} (RNC) loss to obtain a ranked feature \(f_{z_i}\). In practice, we observed that the RNC loss is highly sensitive to the temperature \(\tau\), sometimes leading to \emph{dimensional collapse} \cite{DimensionalCollapse}, where the learned features become overly sparse. To mitigate this, we introduce an \(L_1\) regularization term weighted by \(\lambda\) and train both the encoder and decoder jointly. The modified loss for a single feature, \(l^i_{SupRNC}\), is defined as:
% \begin{equation}
% \begin{aligned}
%     l^i_{SupRNC} = {} & \frac{1}{2N-1} \sum_{\substack{j=1 \\ j\neq i}}^{2N} 
%     \Biggl[-\log\Bigl(
%         \frac{\exp\Bigl(sim(f_{z_i}, f_{z_j})/\tau\Bigr)}
%              {\sum_{f_{z_k} \in S_{i,j}} \exp\Bigl(sim(f_{z_i}, f_{z_k})/\tau\Bigr)}
%     \Bigr)\Biggr] \\
%     & \quad + \lambda\, L1\bigl(x_i, \hat{x}_i\bigr).
% \end{aligned}
% \label{l_SupRNC}
% \end{equation}

\begin{align}
    &l^i_{SupRNC} = {}\\ &\frac{1}{2N-1} \sum_{\substack{j=1 \\ j\neq i}}^{2N} 
    \Biggl[-\log\Bigl(
        \frac{\exp\Bigl(sim(f_{z_i}, f_{z_j})/\tau\Bigr)}
             {\sum_{f_{z_k} \in S_{i,j}} \exp\Bigl(sim(f_{z_i}, f_{z_k})/\tau\Bigr)}
    \Bigr)\Biggr] \\
    & \quad + \lambda\, L1\bigl(x_i, \hat{x}_i\bigr).
\end{align}
\label{l_SupRNC}

Here, \(N\) is the batch size, \(\tau\) controls the sharpness of the similarity distribution, and \(\hat{x}_i\) denotes the ground-truth camera state. The overall loss is:
\begin{equation}
    L_{SupRNC} = \frac{1}{2N}\sum_{i=1}^{2N} l^i_{SupRNC}.
    \label{L_SupRNC}
\end{equation}

By employing the \emph{Supervised Rank-N-Contrast} loss (\(L_{SupRNC}\)), our model learns features that are both highly continuous and systematically ranked by their underlying camera states. This structured organization ensures that visual inputs with similar camera states are closely aligned in the latent space—even when these states originate from different scenes—improving the model’s spatial understanding of the relationship between observations and camera states. Moreover, by regulating the feature distribution, VOCAL generates an interpretable latent space that serves as a critical bridge for integrating with other learning-based models, ultimately increasing the flexibility of visual odometry.

\section{Experiments}
\label{sec:experiment}

\begin{figure*}[t]
  \centering
  % \includegraphics[width=0.8\linewidth]{sec/structure.png}
 \includegraphics[width=0.9\linewidth]{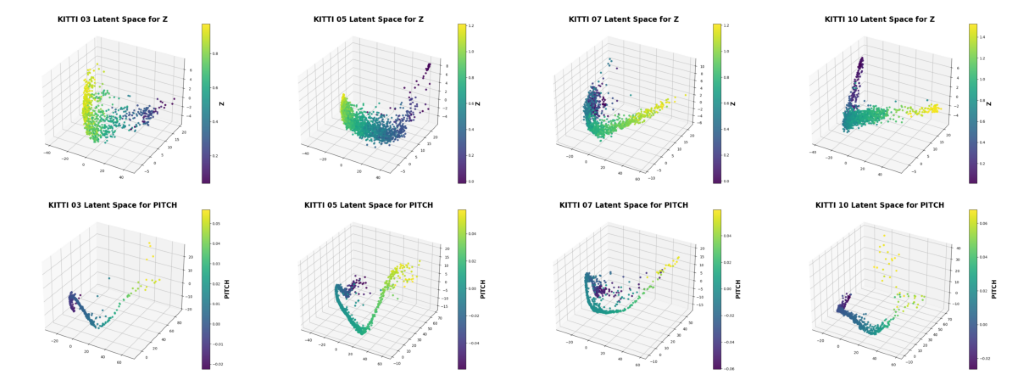}

   \caption{\textbf{Feature Distribution in Latent Space:} Lighter features (yellow) correspond to larger camera motions, while darker features (purple) denote smaller motions. The results, based on KITTI sequences 03, 05, 07, and 10, reveal a continuous gradient from lighter to darker features, highlighting the effective ranking of features according to their camera states.}
   \label{vocal_latent_z_pitch}
\end{figure*}

\paragraph{Implementation Details:}
We represent each camera state using six degrees of freedom (6-DoF): \(\{x, y, z, \mathrm{roll}, \mathrm{pitch}, \mathrm{yaw}\}\). To handle these dimensions, we employ six separate encoder-decoder networks (see Fig.~\ref{vocal_overview}), each responsible for regressing a specific pose component. For the encoder, we use ResNet-18 \cite{ResNet} to process the optical flow between consecutive image pairs and generate a 512-dimensional feature vector \(\mathbf{f} \in \mathbb{R}^{512}\). Each decoder is a three-layer Multi-Layer Perceptron (MLP) that takes \(\mathbf{f}\) as input and predicts one of the six pose parameters.

Prior to training, we compute the optical flow for each pair of consecutive images using the Gunnar-Farneback method. We set the parameters in the OpenCV function as follows: \( \text{pyr\_scale} = 0.5 \), \( \text{levels} = 3 \), \( \text{winsize} = 15 \), \( \text{poly\_n} = 5 \), and \( \text{poly\_sigma} = 1.2 \). The resulting flow maps are then cropped to \( 224 \times 224 \). We apply Gaussian noise augmentation with mean \( \mu = 0.0 \) and standard deviation \( \sigma = 0.05 \).

During training, the \emph{Supervised Rank-N-Contrast} loss (\( L_{\text{SupRNC}} \)) is used to rank encoder-generated features based on their corresponding camera states, resulting in a ranked, interpretable, and continuous latent feature distribution. The decoder then regresses the final pose from this latent representation. For the ranking policy in \( L_{\text{SupRNC}} \), we use the negative \( L_2 \)-norm as the similarity function \( \text{sim}(\cdot,\cdot) \), the \( L_1 \)-norm as the distance function \( d(\cdot,\cdot) \), a temperature factor \( \tau = 2.0 \), and a regularization weight \( \lambda = 2.0 \).

\vspace{-0.5em}

\paragraph{Datasets:}
We conduct our experiments on the KITTI dataset \cite{KITTI}, a standard benchmark for visual odometry and simultaneous localization and mapping. Following \cite{DeepVO}, we train our model on sequences 00, 02, 08, and 09, and evaluate it on sequences 03, 04, 05, 06, 07, and 10.

\vspace{-0.5em}

\paragraph{Evaluation Metrics:}
We evaluate VOCAL’s performance using three primary assessments:
\begin{itemize}
    \item \textbf{Feature Ranking:} We measure how well the ranked features align with their corresponding camera states by computing Spearman’s rank correlation \cite{spearman} and Kendall’s rank correlation \cite{kendall}. Additionally, we visualize the latent-space feature distribution to verify that \(L_{SupRNC}\) effectively ranks features according to camera states.
    \item \textbf{Generalization Capability:} We assess the model’s generalization by training on different fractions of the training set. Specifically, we partition the training data into sizes of 0.2, 0.4, 0.6, 0.8, and 1.0 of the total data, and then measure the Spearman correlation coefficients and VO metrics on the test set.
    \item \textbf{VO Performance:} We follow standard VO evaluation protocols by measuring the average translational RMSE drift (\(\mathrm{t}_{rel}\), in \%) and the average rotational RMSE drift (\(\mathrm{r}_{rel}\), in \(^\circ/100\) m) on trajectory segments of 100–800 m. We compare our results with those of current state-of-the-art learning-based VO methods.
\end{itemize}

\begin{table}

\centering
\small
\resizebox{\columnwidth}{!}{%
\begin{tabular}{|c|cc|cc|cc|}
\hline
Dimension & \multicolumn{2}{c|}{$x$} & \multicolumn{2}{c|}{$y$} & \multicolumn{2}{c|}{$z$}\\
\hline
Correlation & {$r_s \uparrow$} & {$r_k \uparrow$} & {$r_s \uparrow$} & {$r_k \uparrow$} & {$r_s \uparrow$} & {$r_k \uparrow$} \\
\hline
Seq 03 & 0.460 & 0.323 & 0.012 & 0.008 & 0.878 & 0.697 \\
Seq 04 & 0.058 & 0.039 & 0.174 & 0.117 & 0.454 & 0.308 \\
Seq 05 & 0.759 & 0.575 & 0.370 & 0.252 & 0.924 & 0.764 \\
Seq 06 & 0.586 & 0.423 & 0.424 & 0.290 & 0.873 & 0.687 \\
Seq 07 & 0.734 & 0.546 & 0.232 & 0.156 & 0.840 & 0.655 \\
Seq 10 & 0.599 & 0.429 & 0.292 & 0.197 & 0.901 & 0.731 \\
\hline
\end{tabular}
}

\vspace{0.5em} % space between the two sub-tables, optional

\resizebox{\columnwidth}{!}{%
\begin{tabular}{|c|cc|cc|cc|}
\hline
Dimension & \multicolumn{2}{c|}{$roll$} & \multicolumn{2}{c|}{$pitch$} & \multicolumn{2}{c|}{$yaw$}\\
\hline
Correlation & {$r_s \uparrow$} & {$r_k \uparrow$} & {$r_s \uparrow$} & {$r_k \uparrow$} & {$r_s \uparrow$} & {$r_k \uparrow$} \\
\hline
Seq 03 & 0.886 & 0.719 & 0.990 & 0.919 & 0.626 & 0.450 \\
Seq 04 & 0.804 & 0.644 & 0.378 & 0.296 & 0.487 & 0.345\\
Seq 05 & 0.888 & 0.723 & 0.975 & 0.901 & 0.545 & 0.387\\
Seq 06 & 0.860 & 0.687 & 0.936 & 0.816 & 0.467 & 0.339\\
Seq 07 & 0.819 & 0.657 & 0.942 & 0.839 & 0.423 & 0.296\\
Seq 10 & 0.886 & 0.717 & 0.989 & 0.915 & 0.526 & 0.369\\
\hline
\end{tabular}
}

\caption{\textbf{Correlations between Features and Camera States:} High correlation scores for \(z\) translation and \(pitch\) rotation reflect the dominant motion in the KITTI dataset. In contrast, the \(y\) and \(yaw\) dimensions—typically regarded as noise in this dataset—exhibit lower correlation scores for both \(r_s\) and \(r_k\).}

% \caption{Correlation between features and camera states. High correlation scores for \(z\) translation and \(pitch\) rotation reflect the dominance of motion along the \(z\)-axis and changes in \(pitch\) along the \(z\)-axis and changes in \(pitch\) in the KITTI dataset. In contrast, the \(y\) and \(yaw\) dimensions—typically regarded as noise in this dataset—exhibit lower correlation scores for both \(r_s\) and \(r_k\).}

\label{spearman}
\end{table}

\subsection{Feature Ranking}

We analyze the feature distribution produced by our contrastive feature encoder in the latent space. By leveraging the \emph{Supervised Rank-N-Contrast} loss (\(L_{SupRNC}\)), the learned features are expected to closely correlate with the ground-truth camera states. Table~\ref{spearman} reports the Spearman (\(r_s\)) and Kendall (\(r_k\)) rank correlation coefficients, where values approaching 1.0 indicate more effective ranking. Notably, the \(z\)-translation and \(pitch\)-rotation achieve high correlation scores, reflecting the dominant motions along the \(z\)-axis and changes in \(pitch\) in the KITTI dataset. In contrast, the \(y\) and \(yaw\) dimensions, generally considered noise, exhibit lower correlation scores.

Fig.~\ref{vocal_latent_z_pitch} illustrates the feature distribution for \(z\) and \(pitch\) (distributions for the remaining dimensions are provided in the supplementary material \cref{sec:feature}). In the figure, lighter colors (yellow) represent larger camera state values, while darker hues (purple) indicate smaller values. The continuous gradient from light to dark underscores the effective ranking of features by camera state, yielding an interpretable and flexible latent representation that also supports the estimation of the most likely camera states.

\subsection{Generalization Capability}

\begin{figure}[t]
  \centering
  \includegraphics[width=0.85\linewidth]{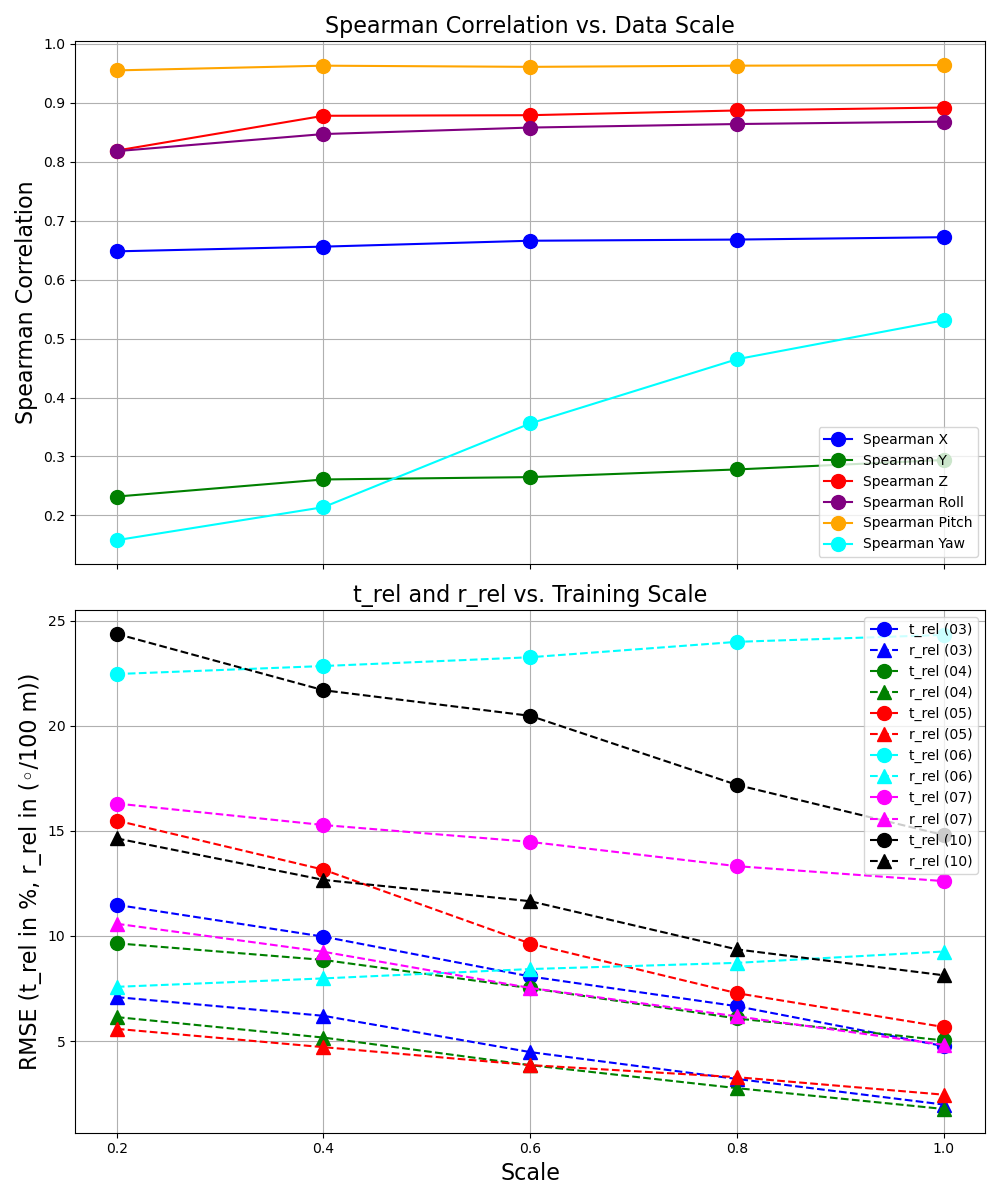}
\caption{
    \textbf{Correlation and VO Metrics vs. Data Scale:} While most learning-based VO methods require extensive training data to achieve high performance, VOCAL converges and delivers competitive results even with limited datasets. As the amount of training data increases, our method achieves higher Spearman rank correlation and lower \(\mathrm{t}_{rel}\) and \(\mathrm{r}_{rel}\) values, indicating that the visual features are effectively ranked according to camera states. This outcome demonstrates VOCAL's flexibility and provides a clearer interpretation of spatial relationships.
}

  \label{vocal_correlation_and_trajectory}
\end{figure}

\begin{table*}[!t]
\centering
\small
\resizebox{\textwidth}{!}{%
\begin{tabular}{|l||c||c|c|c|c|c|c|}
\hline
\multirow{2}{*}{Method} & \multirow{2}{*}{Training dataset} & \multicolumn{6}{c|}{KITTI Sequences (\(\mathrm{t}_{rel}\)/\(\mathrm{r}_{rel}\))} \\
\cline{3-8}
       &                & \textbf{03} & \textbf{04} & \textbf{05} & \textbf{06} & \textbf{07} & \textbf{10} \\
\hline
VISO2-M \cite{VISO2-M}       & {-}               & \underline{8.47}/8.82 & \underline{4.69}/4.49 & 19.22/17.58 & 7.30/6.14 & 23.61/29.11 & 41.56/32.99 \\
SfMLearner \cite{SfMlearner}  & {KITTI 00-08}      & 12.56\(^\ast\)/\underline{4.52}\(^\ast\) & \textbf{4.32}\(^\ast\)/\underline{3.28}\(^\ast\) & 12.99\(^\ast\)/4.66\(^\ast\) & 15.55\(^\ast\)/5.58\(^\ast\) & 12.61\(^\ast\)/6.31\(^\ast\) & 15.25/\underline{4.06} \\

GeoNet \cite{GeoNet}         & {KITTI 00-08}      & 19.41\(^\ast\)/9.80\(^\ast\) & 10.81\(^\ast\)/7.00\(^\ast\) & 22.68\(^\ast\)/7.70\(^\ast\) & 9.90\(^\ast\)/\underline{4.30}\(^\ast\) & 9.82\(^\ast\)/5.90\(^\ast\) & 23.90/9.04 \\
% Depth-VO-Feat \cite{Depth-VO-Feat} & {KITTI 00-08} & 15.76/10.62 & \textbf{3.14}/2.02 & 4.94/\textbf{2.34} & 5.80/2.06 & 6.49/3.56 & 12.45/\underline{3.46} \\
DeepVO \cite{DeepVO}         & {KITTI 00, 02, 08, 09} & 8.49/6.89 & 7.19/6.97 & \textbf{2.62}/\underline{3.61} & 5.42/5.82 & \textbf{3.91}/\underline{4.60} & 8.11/8.83 \\

TartanVO \cite{TartanVO}      & {TartanAir (\(\sim\)40k)} & -         & -         & -          & \textbf{4.72}/\textbf{2.96} & \underline{4.32}/\textbf{3.41} & \textbf{6.89}/\textbf{2.73} \\
VOCAL (ours)                & {KITTI 00, 02, 08, 09} & \textbf{4.60}/\textbf{1.99} & 5.18/\textbf{1.51} & \underline{5.45}/\textbf{2.51} & \underline{5.11}/6.06 & 12.13/4.79 & \underline{7.72}/7.61 \\
\hline
\end{tabular}%
}
\caption{
\textbf{Visual Odometry Results (\(\mathrm{t}_{rel}\)/\(\mathrm{r}_{rel}\)) on KITTI Sequences:} \(\mathrm{t}_{rel}\) denotes the average translational RMSE drift (in \%) over trajectory segments of 100–800 m, and \(\mathrm{r}_{rel}\) denotes the average rotational RMSE drift (in \(^\circ\)/100 m) over trajectory segments of 100–800 m. An asterisk (\(^{\ast}\)) marks training data (which may be overfitted).
}
\label{vocal_vo_comparison}
\end{table*}

We assess VOCAL’s generalization capability by evaluating its performance with varying amounts of training data. Specifically, we partition the training set (sequences 00, 02, 08, and 09) into fractions of the total data (0.2, 0.4, 0.6, 0.8, and 1.0) and evaluate the resulting feature ranking performance on the test set using Spearman’s rank correlation coefficient (\(r_s\)) as well as the VO metrics \(\mathrm{t}_{rel}\) and \(\mathrm{r}_{rel}\).

Fig.~\ref{vocal_correlation_and_trajectory} plots the rank correlation for all six degrees of freedom (DoF) as a function of the training data proportion. Notably, the \(z\) and \(pitch\) dimensions—representing the dominant motions in the KITTI dataset—exhibit gradually increasing \(r_s\) values as more training data becomes available, indicating that our method ranks features more effectively with additional data. In contrast, the \(y\) and \(yaw\) dimensions, which are more susceptible to noise in KITTI, do not converge regardless of the training set size. Additionally, Fig.~\ref{vocal_correlation_and_trajectory} presents the performance of \(\mathrm{t}_{rel}\) and \(\mathrm{r}_{rel}\) on different test datasets, showing that our model performs better as the training data scale increases. Even with a relatively small training set, VOCAL achieves competitive performance compared with other methods (as will be detailed in the next section). This early-stage performance is due to our method’s ability to directly capture the relationship between visual features and camera states through label ranking.

In contrast, many existing learning-based VO models struggle during early training stages and require extensive training to converge due to a lack of interpretability. By comparison, VOCAL not only benefits from larger datasets but also maintains strong performance with limited data, underscoring its potential for effective generalization.

\subsection{VO Performance}
We evaluate VOCAL’s visual odometry performance on the KITTI dataset by training on sequences 00, 02, 08, and 09 and testing on sequences 03–07 and 10. Table~\ref{vocal_vo_comparison} compares translation and rotation accuracy against other VO methods, with the best results in \textbf{bold}, the second-best in \underline{underline}, and an asterisk \((^\ast)\) marking the training data (potentially overfitted). References to the original papers, \cite{DF-VO}, and \cite{LTMVO} are provided for these comparisons.

VISO2-M \cite{VISO2-M}, a classic geometry-based method, is efficient and low-cost but falls short of modern learning-based approaches in accuracy. SfMLearner \cite{SfMlearner} is an end-to-end model for single-view depth and pose estimation. However, it relies heavily on geometric and temporal constraints and performs worse than VOCAL even on its own training data (trained on KITTI sequences 00–08). GeoNet \cite{GeoNet}, also trained unsupervised on sequences 00–08, depends on pretrained depth and flow models; despite these additional resources, it underperforms VOCAL and introduces extra complexity. DeepVO \cite{DeepVO} uses a similar training protocol but depends on an RNN architecture that requires sequential constraints and multiple-frame optimization. In contrast, VOCAL processes a single optical flow input (only two frames) while achieving comparable or superior results. TartanVO shows strong performance on sequences 06 and 07 but demands a large training corpus—hindering reproducibility—and offers limited flexibility for broader integration with learning-based models.

Our experimental results show that VOCAL achieves the best translation performance on sequence 03 and ranks second on sequences 05, 06, and 10. For rotation accuracy, VOCAL attains the best performance on sequences 03, 04, and 05. Notably, although our training data is relatively smaller than that used in \cite{SfMlearner, GeoNet}, VOCAL still performs competitively, even outperforming those methods on training sequences 03–05. These findings demonstrate that VOCAL not only provides an interpretable latent representation but also delivers competitive VO performance.

\section{Conclusion}
\label{sec:conclusion}

We presented \textbf{VOCAL} (Visual Odometry via ContrAstive Learning), a novel framework that reinterprets visual odometry as a label-ranking problem by integrating Bayesian inference with contrastive representation learning. By organizing visual features in a latent space according to underlying camera states, VOCAL produces an interpretable, continuous representation that effectively captures spatial relationships.

Our experiments on the KITTI dataset show that VOCAL achieves competitive translation and rotation accuracy compared to other learning-based VO methods, while avoiding reliance on geometric or temporal constraints. Unlike conventional approaches that depend on handcrafted features or sequential inputs, VOCAL operates on optical flow from only two frames, enabling a flexible and efficient model that generalizes well across different scenarios.

Moreover, VOCAL serves as a bridge between visual odometry and other learning-based models. Traditional Bundle-Adjustment–based methods offer little control over the latent space, and their optimization logic rarely generalizes to other learning-based frameworks. In contrast, our approach produces a visualized, interpretable, and ranked latent representation that can be seamlessly integrated with additional sensor modalities and learning-based systems, thereby enhancing performance in multimodal applications.

Beyond visual odometry, VOCAL offers a new perspective on learning spatial and conceptual relationships across modalities. The label-ranking principle in latent space can be extended to various space-to-space or concept-to-concept tasks, such as aligning visual inputs with camera states, associating 2D images with text or 3D reconstructions, or linking sensor data (e.g., depth, IMU, GPS) with high-level representations (e.g., language, multi-agent states, spatial concepts). Compared to classical graph-based approaches, our framework provides new insights and broader potential for advancing spatial intelligence.
  
In summary, VOCAL advances the state of visual odometry, provides a foundation for flexible and interpretable multimodal systems, and opens new directions for solving broader space-to-space problems. Future work will focus on incorporating additional sensor modalities, refining latent representations through advanced learning strategies, and validating the approach in more diverse real-world and conceptual domains. We believe the ideas behind VOCAL have the potential to drive significant progress in spatial intelligence—and beyond.

% \input{sec/6_acknowledgement}
%reviewing is double blind. 
{
    \small
    \bibliographystyle{ieeenat_fullname}
    \bibliography{main}
}
\clearpage
\setcounter{page}{1}
\maketitlesupplementary

\section{Resources}
For additional resources and demo videos, please visit:
\href{https://github.com/huang-chiyao/vocal}{https://github.com/huang-chiyao/vocal}.

\section{Feature Distribution}
\label{sec:feature}
In this section, we present the detailed feature distribution for all six degrees of freedom across the test datasets (KITTI sequences 03–07 and 10).

\begin{figure}[h]
  \centering
  \includegraphics[width=0.9\linewidth]{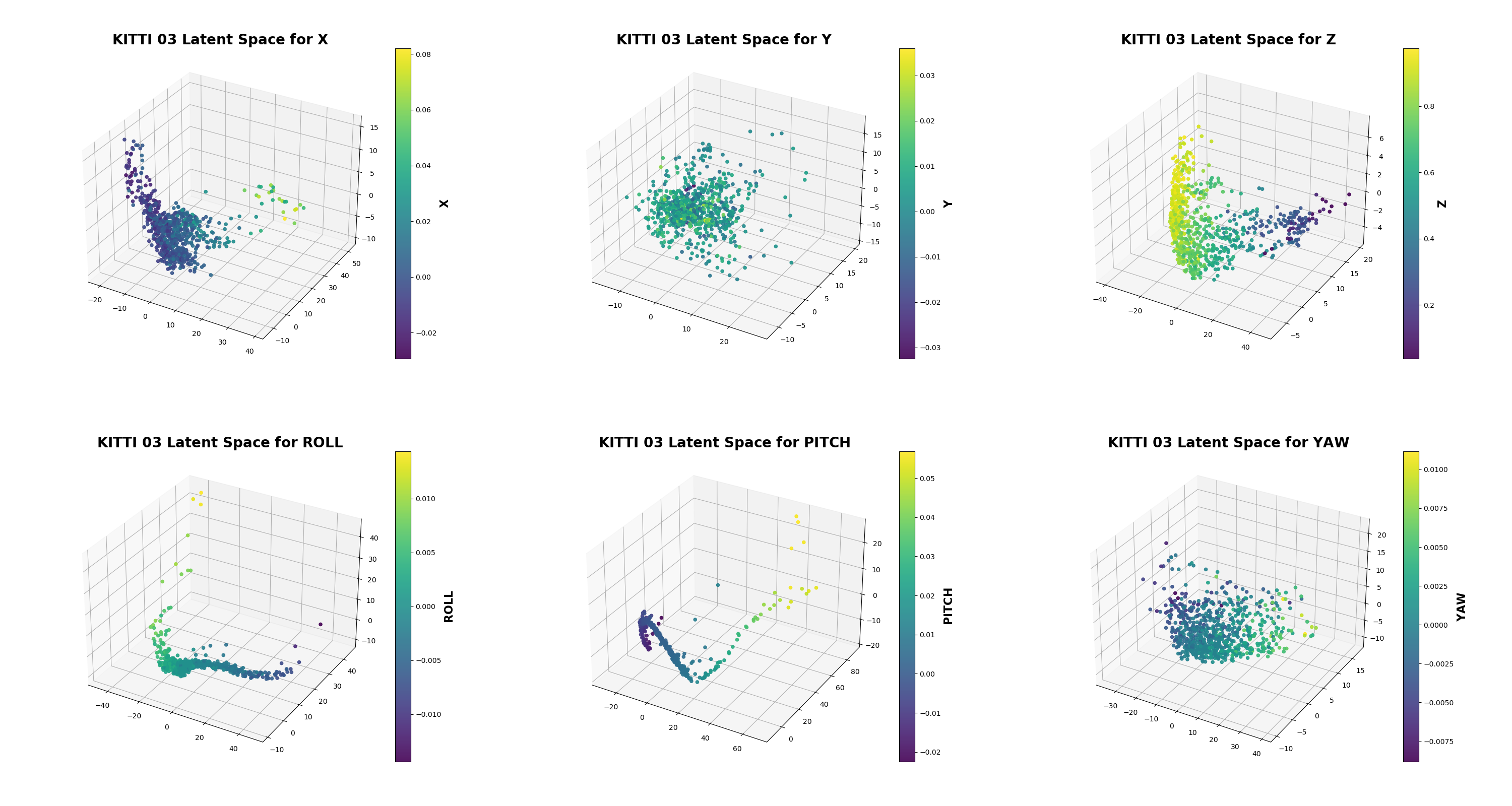}

   \caption{KITTI 03 Feature Distribution}
   \label{vocal_latent_3D_combined_03}
\end{figure}

\begin{figure}[h]
  \centering
  \includegraphics[width=0.9\linewidth]{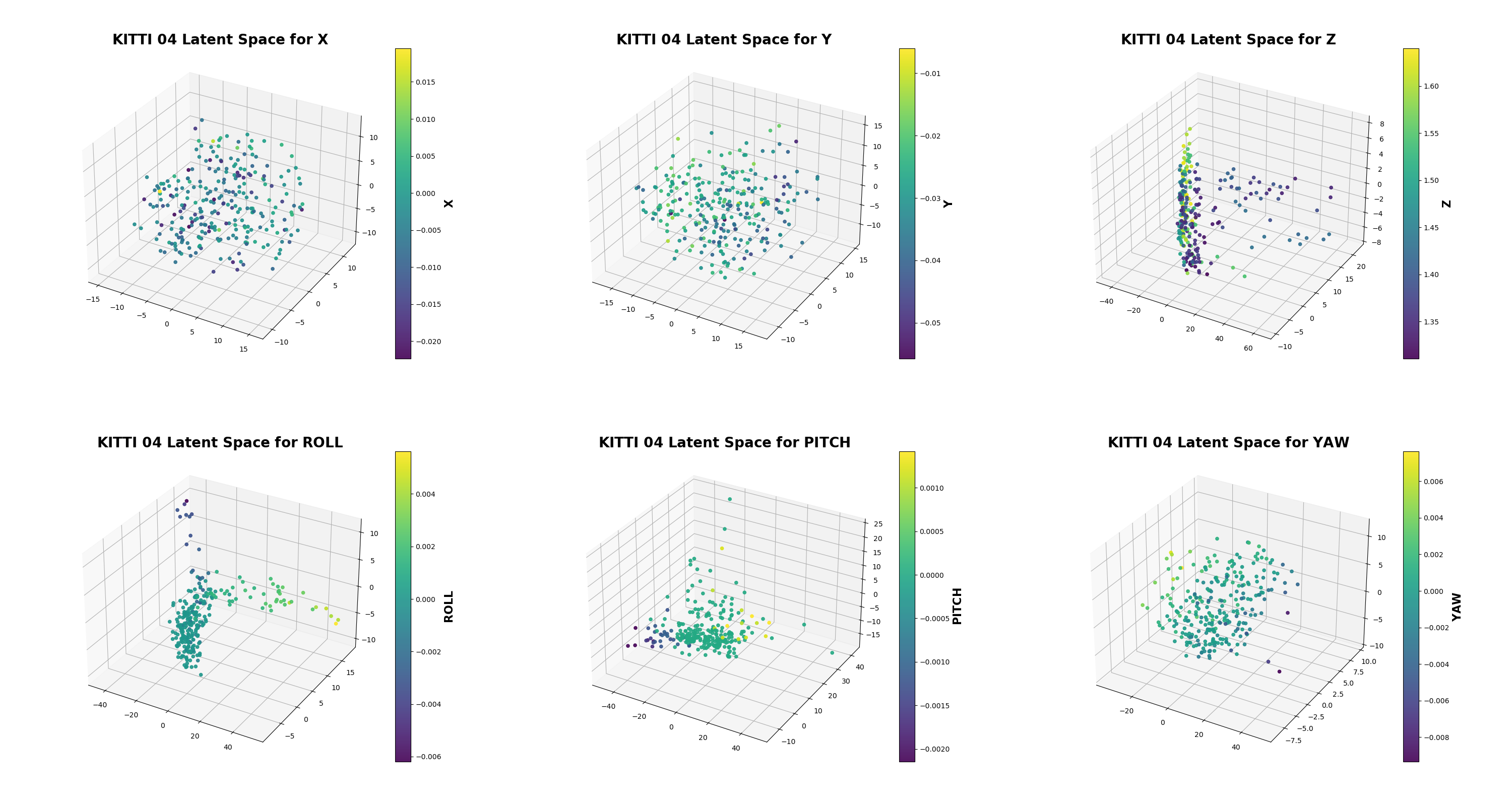}

   \caption{KITTI 04 Feature Distribution}
   \label{vocal_latent_3D_combined_04}
\end{figure}

\begin{figure}[h]
  \centering
  \includegraphics[width=0.9\linewidth]{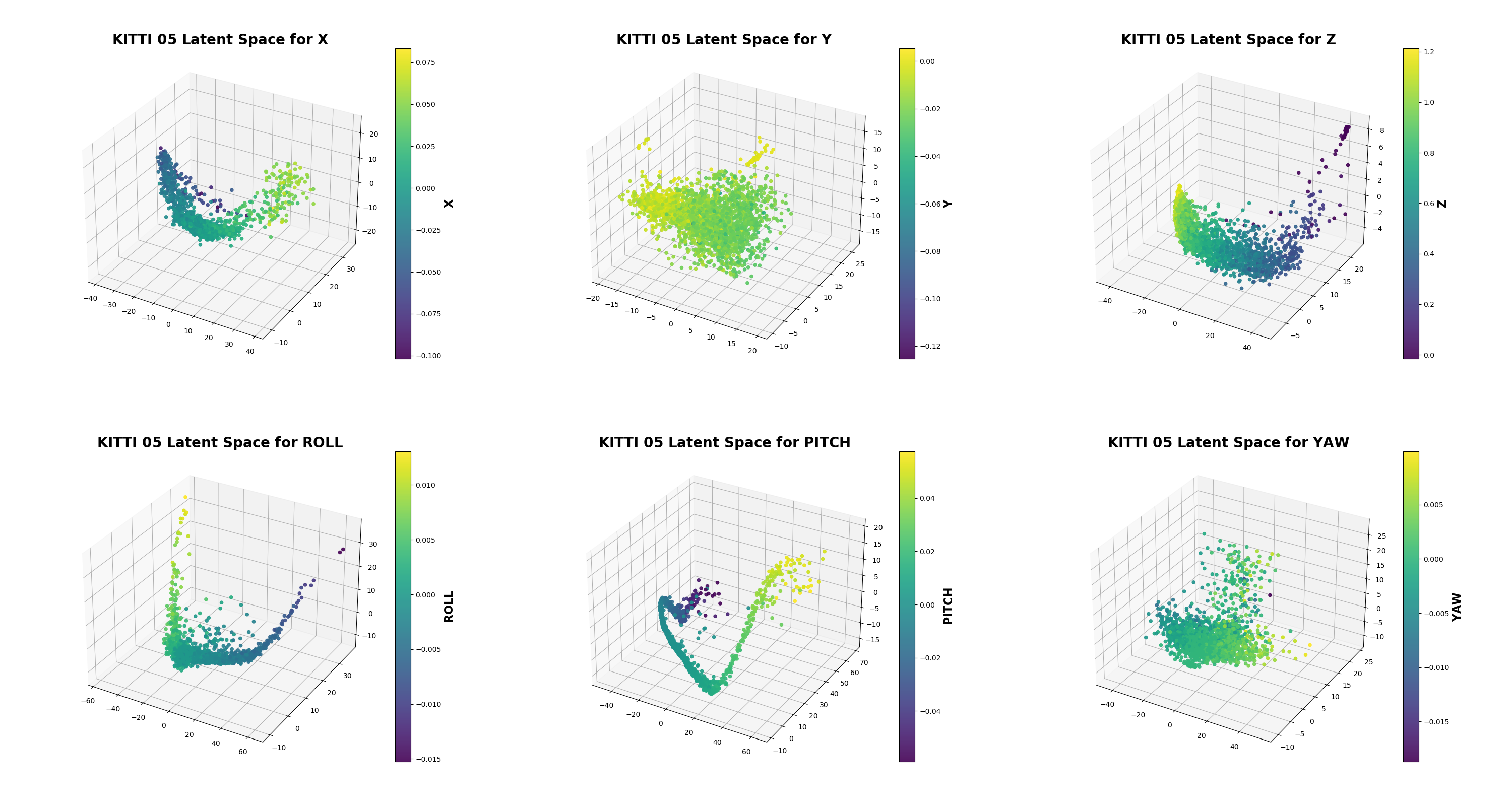}

   \caption{KITTI 05 Feature Distribution}
   \label{vocal_latent_3D_combined_05}
\end{figure}

\begin{figure}[h]
  \centering
  \includegraphics[width=0.9\linewidth]{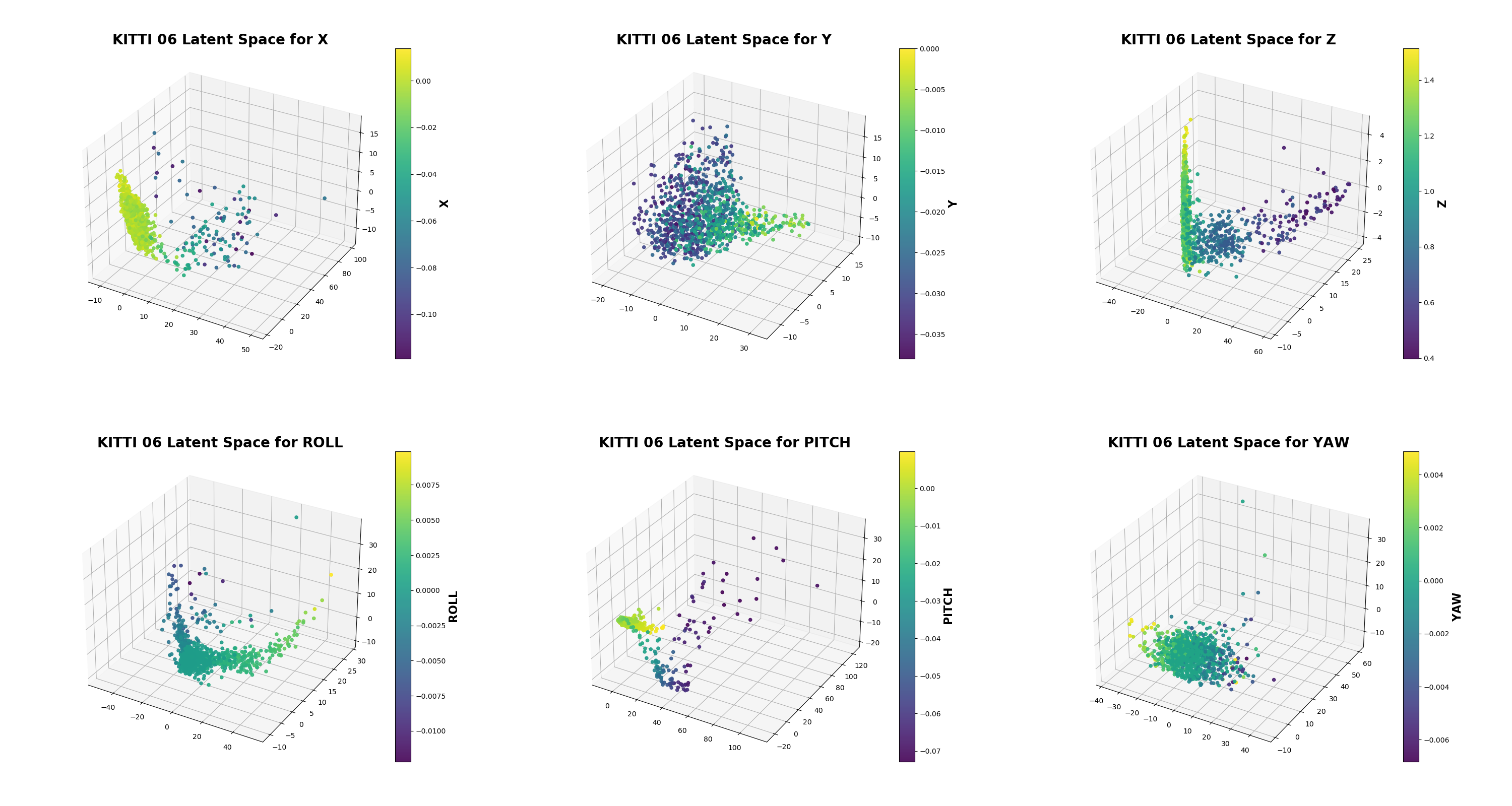}

   \caption{KITTI 06 Feature Distribution}
   \label{vocal_latent_3D_combined_06}
\end{figure}

\begin{figure}[h]
  \centering
  \includegraphics[width=0.9\linewidth]{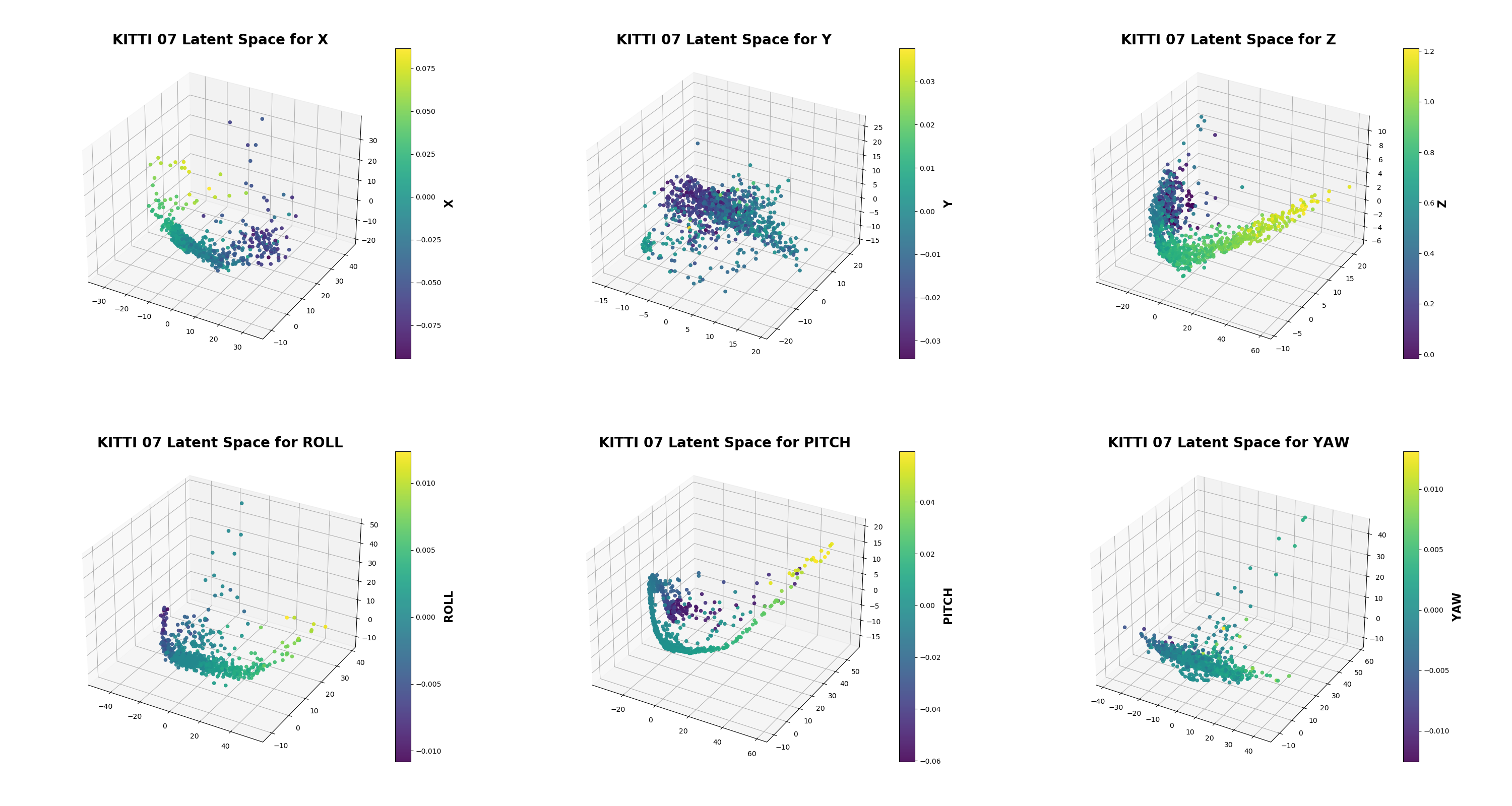}

   \caption{KITTI 07 Feature Distribution}
   \label{vocal_latent_3D_combined_07}
\end{figure}

\begin{figure}[h]
  \centering
  \includegraphics[width=0.9\linewidth]{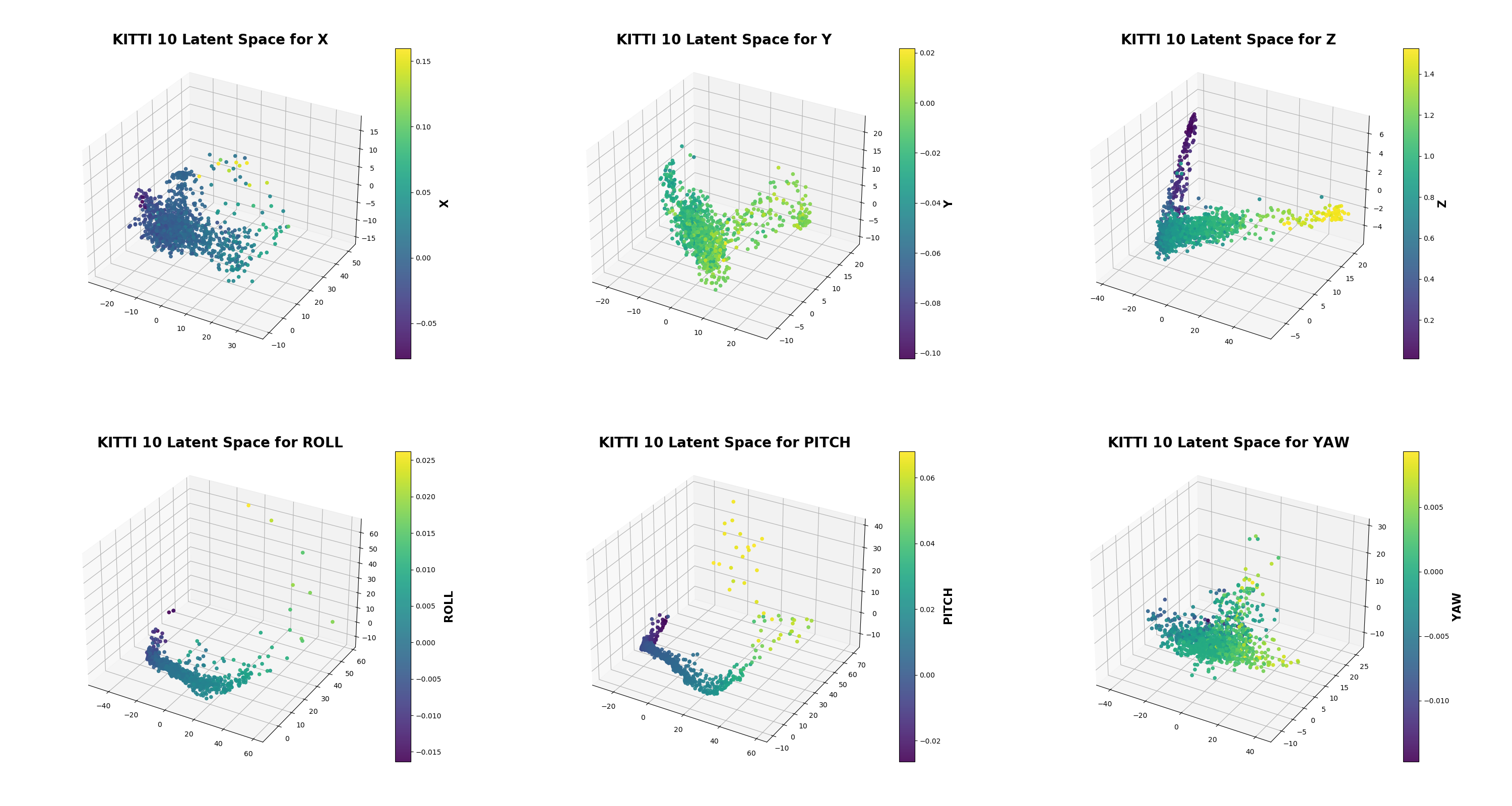}

   \caption{KITTI 10 Feature Distribution}
   \label{vocal_latent_3D_combined_10}
\end{figure}

\clearpage

% \section{Trajectory}
% \label{sec:trajectory}
% In this section, we present the 3D trajectories for the test datasets (KITTI sequences 03–07 and 10).

% \begin{figure}[h]
%   \centering
%   \includegraphics[width=0.75\linewidth]{fig/trajectory/vocal_plot_3d_03.png}

%    \caption{KITTI 03 Trajectory}
%    \label{vocal_plot_3d_03}
% \end{figure}

% \begin{figure}[h]
%   \centering
%   \includegraphics[width=0.75\linewidth]{fig/trajectory/vocal_plot_3d_04.png}

%    \caption{KITTI 04 Trajectory}
%    \label{vocal_plot_3d_04}
% \end{figure}

% \begin{figure}[h]
%   \centering
%   \includegraphics[width=0.75\linewidth]{fig/trajectory/vocal_plot_3d_05.png}

%    \caption{KITTI 05 Trajectory}
%    \label{vocal_plot_3d_05}
% \end{figure}

% \begin{figure}[h]
%   \centering
%   \includegraphics[width=0.75\linewidth]{fig/trajectory/vocal_plot_3d_06.png}

%    \caption{KITTI 06 Trajectory}
%    \label{vocal_plot_3d_06}
% \end{figure}

% \begin{figure}[h]
%   \centering
%   \includegraphics[width=0.75\linewidth]{fig/trajectory/vocal_plot_3d_07.png}

%    \caption{KITTI 07 Trajectory}
%    \label{vocal_plot_3d_07}
% \end{figure}

% \begin{figure}[h]
%   \centering
%   \includegraphics[width=0.75\linewidth]{fig/trajectory/vocal_plot_3d_10.png}

%    \caption{KITTI 10 Trajectory}
%    \label{vocal_plot_3d_10}
% \end{figure}

\end{document}